\pdfoutput=1
\PassOptionsToPackage{table}{xcolor}
\documentclass[11pt]{article}

\usepackage{acl}

\usepackage{times}
\usepackage{latexsym}
\usepackage{amsmath}
\usepackage[table]{xcolor} 
\usepackage{booktabs}  

\usepackage[T1]{fontenc}

\usepackage[utf8]{inputenc}

\usepackage{microtype}

\usepackage{inconsolata}

\usepackage{graphicx}

\title{Boolean-aware Attention for Dense Retrieval}

\author{
 \textbf{Quan Mai\textsuperscript{1}},
 \textbf{Susan Gauch\textsuperscript{1}},
 \textbf{Douglas Adams\textsuperscript{2}},
\\
 \textsuperscript{1}Department of Electric Engineering and Computer Science,\\
 \textsuperscript{2}Department of Sociology and Criminology, \\
 University of Arkansas
\\
 \texttt{quanmai, sgauch, djadams @uark.edu}
}

\begin{document}
\maketitle
\begin{abstract}
We present Boolean-aware attention, a novel attention mechanism that dynamically adjusts token focus based on Boolean operators (e.g., ``and", ``not"). Our model employs specialized Boolean experts, each tailored to amplify or suppress attention for operator-specific contexts. A predefined gating mechanism activates the corresponding experts based on the detected Boolean type. Experiments on two Boolean retrieval datasets demonstrate that integrating \emph{BoolAttn} with BERT greatly enhances the model's capability to process Boolean queries.
\end{abstract}

\section{Introduction}
Boolean operations such as \emph{and}, \emph{or}, and \emph{not} are fundamental components of human logic and reasoning, serving as essential tools for combining, including, or excluding specific information in queries. These operations are particularly valuable when users have complex and precise information needs, such as “Movies set in Vietnam but \emph{not} about war” or “Animals found only in Brazil \emph{and} Mexico.” However, traditional lexical matching methods like BM25 \citep{robertson2009probabilistic} often fail to adequately handle queries involving exclusions, frequently retrieving results that include undesired information. Similarly, transformer-based architectures such as BERT \citep{devlin2018bert} encounter challenges in processing logical operations \citep{mai2024setbert, malaviya2023quest}. These models lack explicit mechanisms to interpret Boolean cues and struggle with contextual dependencies, leading to suboptimal performance in handling logical semantics. To overcome these limitations, we present a Boolean-aware attention mechanism, an extension of the standard attention mechanism, designed to explicitly model Boolean logic with enhanced sensitivity to logical cues and their contextual scopes. Our method ensures that models can better capture and process logical constructs, enabling improved performance on tasks involving complex Boolean queries.

\section{Boolean-aware attention}
\begin{figure}[h]
  \includegraphics[width=\columnwidth]{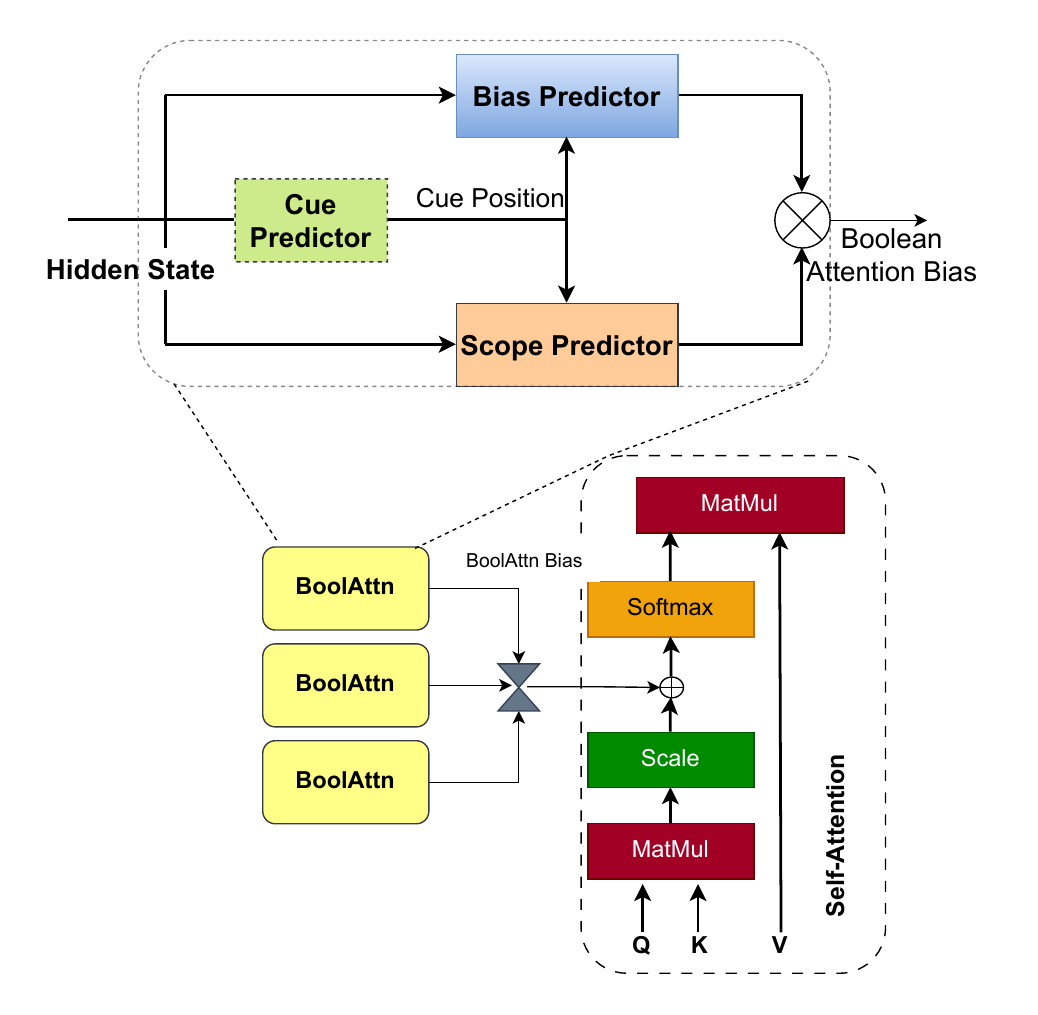}
  \caption{Boolean-aware Attention architecture.}
  \label{fig:boolattn}
\end{figure}

Our Boolean-aware attention mechanism (or, \emph{BoolAttn}, for short) consists of four key components: (1) a \textbf{Cue Predictor} (\ref{sec: cuepred}) that detects Boolean cues within the input sequence, (2) a \textbf{Contextual Scope Predictor} (\ref{sec: scopepred}) that determines the scope of Boolean operators, ensuring their influence is properly localized, (3) a \textbf{Bias Predictor} (\ref{sec: biaspred})  that adjusts attention scores based on Boolean operations using operator-specific embeddings and positional dependencies, and (4) a \textbf{modified attention mechanism} incorporating Boolean-specific biases into the attention scores.  By explicitly modeling Boolean logic while preserving contextual dependencies, \emph{BoolAttn} enables transformers to better interpret logical structures in natural language queries. Furthermore, \emph{BoolAttn} is lightweight, modular, and fully compatible with existing transformer architectures, serving as a \textit{plug-and-play} enhancement for self-attention layers. The overall architecture of \emph{BoolAttn} is illustrated in Figure~\ref{fig:boolattn}. 

\begin{figure}[t]
  \includegraphics[width=\columnwidth]{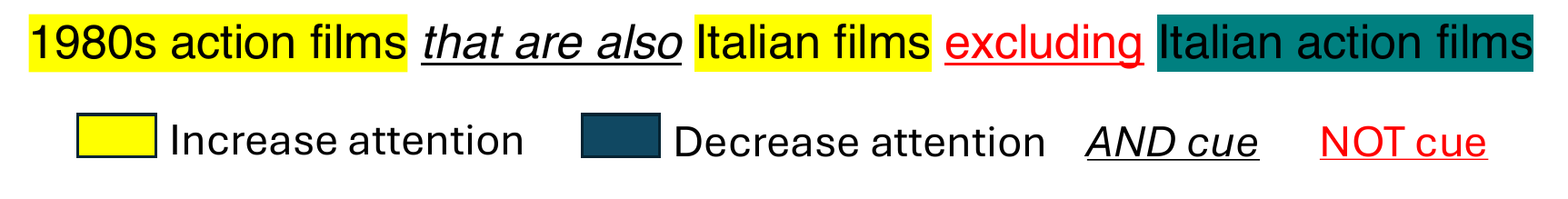}
  \caption{Token-wise attention scores are adjusted based on Boolean type. Tokens affected by \emph{and} will receive more attention while tokens affected by \emph{not} will receive less attention score.}
  \label{fig:demonstration}
\end{figure}
\paragraph{Boolean operator embeddings} We use operator-specific embeddings $\mathbf{o}\in\mathbf{R}^{d_o}$ to encode the unique semantic roles of Boolean operators, such as the negation effect of \emph{not} or the conjunctive nature of \emph{and}. These embeddings guide other modules to produce operator-specific outputs, enabling the model to adapt to the logical properties of each operator.

\subsection{Boolean Cue Predictor} \label{sec: cuepred}
Boolean cues play a pivotal role in determining the logical structure of queries. These cues dictate the scope and interaction between different query components, thereby influencing the embedding of the query. For instance, cues for \emph{not} (negation) include keywords like “not,” “excluding,” and “without,” while cues for \emph{and} include “and,” “as well as,” “also,” and “including.” The Cue Predictor (CuePred) is designed to identify the positions of Boolean cues within the input sequence. Since the positional information of cues remains invariant for a given input sequence, the CuePred is employed exclusively in the first transformer layer. The detected cue positions are then propagated through subsequent layers to inform the Scope Predictor and Bias Predictor at each corresponding layer. Notably, Boolean-related retrieval datasets \citep{malaviya2023quest} do not explicitly annotate Boolean cues. Therefore, cue detection is formulated as an auxiliary task during the pretraining phase, enabling the model to learn cue positions implicitly and generalize effectively to downstream tasks.

The operator embeddings $\mathbf{o}$ are projected into the model's hidden space using a linear transformation
$$\mathbf{o'} = \mathbf{W_o}\mathbf{o} + \mathbf{b}_o$$
where $\mathbf{W}_o \in \mathbf{R}^{d_h\times d_o}$ and $\mathbf{b}_o \in \mathbf{R}^{d_h}$ are learnable parameters, $d_h$ and $d_o$ denote the hidden and operator embedding dimensions, respectively. The projected embedding is then added to the hidden states $H$ before going through a shared linear layer to compute cue scores, followed by sigmoid activation function to compute the cue probabilities:
$$p_c = \sigma(\mathbf{W}_c(\mathbf{H+\mathbf{o'}) + \mathbf{b}_c)}$$
where $\mathbf{W}_c\in \mathbf{R}^{1\times d_h}$ and $\mathbf{b}_c\in \mathbf{R}$ are learnable parameters. Cue probabilities $p_c\in\mathbf{R}^{B\times L}$ indicates the likelihood of each position being a cue.

\subsection{Boolean Scope Predictor}\label{sec: scopepred}
The Scope Predictor (ScopePred) identifies the affected span of tokens (the \emph{scope}) for a given Boolean operator. For example, in the sentence shown in Figure~\ref{fig:demonstration}, the scope of \emph{not} is ``Italian action film," and the model downweights the importance of these tokens in the \emph{BoolAttn} mechanism. The predictor leverages local contextual features and conditions on operator embeddings to determine the scope. It combines a Conv1D layer to capture local dependencies and a FiLM layer (Feature-wise Linear Modulation, \citealp{perez2018film}) to specialize outputs for each operator.

Boolean operators primarily influence tokens near the cue, as observed in phrases ``not happy," where negation affects the adjacent token. To effectively capture such local dependencies, ScopePred utilizes a Conv1D layer. Especially, the hidden states $H \in \mathbf{R}^{B\times L\times d_h}$ and cue probabilities $p_c\in\mathbf{R}^{B\times L}$ are concatenated and subsequently passed through a Conv1D layer:

$$S = \mbox{Conv1D}([\mathbf{H; \mathbf{p}_c}])$$
This layer efficiently captures local patterns and dependencies around the cue, making it particularly suitable for modeling the impact of Boolean operators on nearby tokens. To adapt the predictions for each Boolean operator, the FiLM layer modulates the Conv1D output using operator embeddings $\mathbf{o}$. Scale and shift factors are computed as:
$$\gamma=\mathbf{W}_{\gamma}o + \mathbf{b}_{\gamma}\mbox{, }\mbox{ }\mbox{ } \beta=\mathbf{W}_{\beta}o + \mathbf{b}_{\beta}$$
These factors are then applied to modulate the Conv1D output:
$$S = \gamma S + \beta$$
The modulated logits $S$ are passed through a sigmoid activation to obtain scope probabilities:
$$P_{scope} = \sigma(S)$$
Since the scope is binary (0 for outside scope, 1 for inside scope), we use the Gumbel-Sigmoid trick to enable discrete scope sampling while preserving differentiability. The scope is determined by applying a learnable threshold $\theta$ to the probabilities $P$ obtained from the Gumbel-Sigmoid:
\begin{equation}
\label{eq:scope}
    \mbox{S} = \mathbf{I}(P > \theta)
\end{equation}
To allow gradient flow through the non-differentiable thresholding, we use a Straight-Through Estimator (STE). This approach ensures that the model learns to make discrete scope predictions effectively.

Different Boolean operators impose unique structural constraints on token interactions, necessitating specialized mechanisms for scope prediction. The \emph{and} operator establishes bidirectional reinforcement between operands (or entities), ensuring mutual influence, whereas \emph{not} applies a unidirectional effect, modifying the meaning of specific tokens within its scope. When a negation cue like "not" appears, it alters the interpretation of a single token or phrase rather than forming a bidirectional relationship. Conversely, \emph{and} enforces a reciprocal dependency—e.g., in ``A and B," both A and B should influence each other—requiring a pairwise rather than token-wise scope. To capture these differences, we define the scope matrices for \emph{and} and \emph{or} as follows:

$$S_{mutual} = S \cdot S^T $$

where $S_{mutual} \in \mathbf{R}^{B \times heads\ \times L \times L}$ captures symmetric token influence, while \emph{not} retains a unary mask of shape $(B, heads, L, 1)$ as it requires no pairwise expansion.

\subsection{Bias Predictor}\label{sec: biaspred}
We use a Boolean-aware bias module  to model token-level biases in \emph{BoolAttn} by incorporating positional dependencies and operator-specific embeddings. This module learns to adjust attention bias based on the relative positions of Boolean cue (from the CuePred) and the surrounding context.
\paragraph{Relative position} To capture the structural influence of Boolean operators, the model first computes the relative positions of tokens with respect to the Boolean cues (e.g., ``without"). Given a sequence of length $L$, the relative position of token $i$ with respect to a Boolean cue at position $c$ is defined as:
$$r_i = i - c$$
For the negation operator (\emph{not}), its influence is asymmetric, affecting only subsequent tokens. To ensure this, we clip negative relative postions:
$$r_i = \max(0, i - c), \mbox{    for \emph{not}}$$
For conjunctions (\emph{and}) and disjuntions (\emph{or}), influence is bidirectional, meaning the relative positions are clamped with a symmetric window $\pm d$:
$$r_i = \mbox{clip}(i-c, -d, d), \mbox{    for \emph{and} /\emph{or}}$$
To assign different levels of importance based on distance from the cue, we can use position embeddings. To make this module lightweight, instead of using learnable positional embeddings,  we use a Gaussian kernel to weight positions. The idea of using Gaussian kernel is we hypothesize that tokens closer to the cue are more likely to be influenced:
$$w(r_i)=\exp \left(-\frac{r_i^2}{2*\sigma^2}\right)$$
where $\sigma$ is a learnable parameter controlling the spread of the Gaussian influence. The computed positional weights $w(r)$ are concatenated with token hidden state $H$ to form an enriched representation before passed through a Feed-Forward Network (FFN) to generate context-aware bias scores:
$$\mathbf{b} = \mbox{FFN}([\mathbf{H}, w(r)])$$
To account for the distinct biases associated with different Boolean operations, an operator-specific gate dynamically modulates the computed biases:
\begin{equation}
\label{eq:bias}
\mathbf{b'} = \mathbf{b}\cdot \sigma(\mathbf{W}_{op}\mathbf{e}_{op})
\end{equation}
The final bias values are regularized using softplus to explicitly ensure non-negativity.
\subsection{Operator-gating mechanism}
After having Scope (Eq.~\ref{eq:scope}) and Bias (Eq.~\ref{eq:bias}) of each Boolean operator, we integrate the Boolean reasoning into the transformer's attention mechanism. This integration employs a gating mechanism, dynamically weighting each operator's influence based on its presence in the input.

The attention bias contribution for a Boolean operator \emph{op} is formulated as:
$$S_{op} = G_{op} \cdot \mbox{Scope}_{op} \cdot \mbox{Bias}_{op}$$
where $G_{op}\in \{0;1\}$ is a gating variable that indicates the presence of the operator in the input sequence. This gate can be either learned through an experts classifier or provided as an auxiliary label, enabling conditional application of the attention bias based on the occurrence of the operator.
Since we aim to reinforce tokens of \emph{and}, \emph{or} and downweight the importance of tokens affected by \emph{not} (Appendix \ref{ap:whydownweight}), the final Boolean-aware attention bias is then formulated as:
$$S_{Boolean} = S_{and} + S_{or} - S_{not}$$
The Boolean-aware bias is added to the raw attention scores, ensuring that the logical structure of sentences influences the attention mechanism:
$$\mbox{Attention Score} = \mbox{Softmax}(\frac{\mathbf{Q}\mathbf{K}^T}{\sqrt{d}} + S_{Boolean})\mathbf{V}$$
where $\mathbf{Q}, \mathbf{K}, \mathbf{V}$ are original attention matrices in the self-attention mechanism \citep{vaswani2017attention}.



\section{Evaluation}
We evaluate \emph{BoolAttn} by integrating it into BERT (\textit{uncased} configuration), forming \emph{Bool-BERT}, and testing it on two Boolean retrieval datasets: Quest~\cite{malaviya2023quest} and BoolQuestions~\cite{zhang2024boolquestions}. Prior to the main evaluation, we pretrain \emph{BoolAttn} on the GPT-generated dataset by \citealp{mai2024setbert} to ensure effective initialization (see Appendix~\ref{ap:pretrain}). All experiments are repeated three times with different random seeds \{0, 42, 1234\}, and we report the average results.

\begin{figure}
    \centering
    \includegraphics[width=\columnwidth]{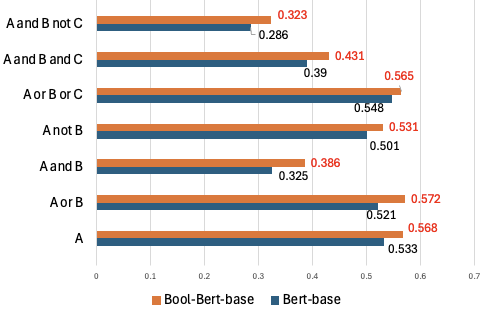}
    \caption{Recall@100 across different query templates.}
    \label{fig:templates}
\end{figure}

\paragraph{Entity-seeking queries}
QUEST features queries with implicit Boolean operations, testing models' ability to understand logical relationships among entities. We employ Bool-BERT as the query encoder in a dual-encoder setup, paired with BERT as the document encoder. Retrieval performance is measured using average Recall@K, with K set to $\{20,50,100,1000\}$, as shown in Table~\ref{tab:recall}. Additionally, we assess the retrievers on seven distinct query structures to capture nuances in retrieval capabilities, with results presented in Figure~\ref{fig:templates}. These findings demonstrate that integrating the \emph{BoolAttn} plugin enhances BERTs' effectiveness in Boolean query understanding.

        
        
\begin{table}[]
\begin{tabular}{lcccc}
\hline
                           & \multicolumn{4}{c}{Avg. Recall@K}                                 \\ \cline{2-5} 
Retriever                  & 20             & 50             & 100            & 1000           \\ \hline
BERT\textit{-base}                  & 0.228          &0.344          & 0.443          & 0.760          \\
\rowcolor{green!40}
Bool-BERT\textit{-base} & 0.254          & 0.379          & 0.483         & 0.788 \\ \hline
BERT\textit{-large }                & 0.238          & 0.364          & 0.467          & 0.792          \\
\rowcolor{red!30}
Bool-BERT-\textit{large}                    &  0.266         & 0.395          & 0.495          & 0.812          \\
\bottomrule
\end{tabular}
\caption{Bool-BERT retrievers outperform standard BERT-based models in retrieval tasks.}
\label{tab:recall}
\end{table}

\paragraph{BoolQuestions} We use the best-performing models from fine-tuning on Quest as starting checkpoints for continuous learning on BoolQuestions' supplementary training dataset, focusing on \emph{and}, \emph{or}, and \emph{not} operations. Evaluation is conducted on the corresponding test set, adhering to the original experimental setup by reporting Mean Reciprocal Rank at top-10 (MRR@10). The results in Table~\ref{tab:boolquestions} highlight the effectiveness of \emph{BoolAttn} in enhancing performance for Boolean-related QA tasks. For comparison, the results of \textit{distillbert-dot-v5} are taken from the original paper.
\begin{table}[]
\centering
\begin{tabular}{llll}
\hline
                   & \multicolumn{3}{c}{MRR@10} \\ \cline{2-4} 
                   & AND     & OR      & NOT    \\ \hline
distillbert-dot-v5 & 0.399   & 0.530   & 0.130   \\
\rowcolor{yellow!20}
BERT\textit{-base}          & 0.412   & 0.545   & 0.128  \\
\rowcolor{blue!20}
Bool-BERT\textit{-base}           & 0.480   & 0.586   & 0.165  \\
\rowcolor{red!20}
Bool-BERT\textit{-large}           & 0.532   & 0.610   & 0.182  \\
\bottomrule
\end{tabular}
\caption{Performance on MS MARCO's BoolQuestions.}
\label{tab:boolquestions}
\end{table}
\section{Conclusion}
In this work, we introduced \emph{BoolAttn}, a novel attention mechanism that dynamically adjusts attention weights based on Boolean operators, effectively capturing the distinct structural influences of conjunctions, disjunctions, and negations. By integrating \emph{BoolAttn} into BERT, we greatly improved performance on Boolean retrieval tasks, demonstrating its effectiveness in processing complex logical relationships within queries. These results highlight the potential of Boolean-aware attention mechanisms to enhance transformers' ability to understand and handle nuanced Boolean queries.

\newpage
\section*{Limitations}
While \emph{BoolAttn} greatly enhances the performance of the backbone BERT model and is designed to be lightweight, it still introduces a considerable number of parameters. Specifically, Bool-BERT\textit{-base} has approximately 121M parameters, while the \textit{large} version reaches around 378M parameters. This indicates that \emph{BoolAttn} adds roughly 10\% more parameters to the underlying transformer architecture, increasing \emph{computation cost and latency}. To mitigate this overhead, future work could explore parameter-sharing strategies across Boolean modules or investigate low-rank factorization techniques to reduce the parameter count without sacrificing performance.


Moreover, the Scope and Bias Predictors depend on accurate inputs from the Cue Predictor, making the system vulnerable to error propagation. If cues are misidentified, scope and bias predictions may be incorrect. This is particularly challenging when Boolean cues are implied rather than explicit. For example, in the query ``Find articles on climate policy favoring renewables over fossil fuels," the comparative phrase implicitly negates fossil fuels without an explicit \emph{not} operator. If the Cue Predictor misses this, the system might retrieve documents about both ``renewables" and ``fossil fuels", misinterpreting the intended negation. This could lead to misleading search results, especially in \emph{sensitive domains} like legal, medical, or financial information retrieval.

Additionally, the absence of explicit ground truth for scope labels in existing Boolean retrieval datasets presents a significant challenge. Since no direct supervision is available for scope boundaries, the optimization of \emph{BoolAttn} relies on indirect signals from retrieval performance, which may not fully capture the intended scope dynamics. This lack of supervision makes it difficult to ensure that \emph{BoolAttn} learns the desired scope representation, potentially leading to suboptimal generalization.

Lastly, the limited size of Boolean retrieval datasets constrains the scalability of \emph{BoolAttn}. Although the model is effective in enhancing Boolean query understanding, the small training corpus limits its potential to learn complex linguistic patterns and generalize to diverse scenarios. This \emph{scalability issue} poses challenges for deploying \emph{BoolAttn} in large-scale retrieval systems, as its performance may degrade without sufficient training data.

\section*{Ethical considerations}
This work utilizes Quest, BoolQuestions, and GPT-generated datasets strictly for research purposes, consistent with their original licensing terms. The \emph{BoolAttn} module is designed for academic evaluation of Boolean retrieval models. Any derivative models or results are intended solely for research and should not be used outside of academic contexts. We ensured that the datasets used in this study, including QUEST and BoolQuestions, did not contain personally identifiable information (PII) or offensive content. A thorough review was conducted to confirm that the data complied with ethical standards.

\bibliography{acl_latex}

\appendix

\section{Why downweight attention scores for negation}
\label{ap:whydownweight}
Negation treatment is a challenging problem for current dense retrieval models \cite{zhang2024boolquestions, malaviya2023quest, mai2024setbert} and vision-language retrieval models \cite{bui2024nein}.

The phenomenon where attempting to suppress specific thoughts leads to their increased prominence is well-documented in psychological literature. This is exemplified by the ``pink elephant" paradox: when instructed not to think of a pink elephant, ones often find it challenging to avoid the image. This counterproductive effect is known as ironic process theory \cite{wegner1994ironic}, which posits that deliberate efforts to suppress certain thoughts can make them more persistent. 

In the context of language processing, negation requires the brain to first represent the concept being negated before it can suppress or modify it. This two-step process suggests that negation involves initial activation followed by inhibition \citep{dudschig2021processing}.

Building upon these insights, our hypothesis proposes that in Boolean-aware attention mechanisms, tokens affected by negation should be downweighted. By reducing the influence of these negated tokens, the model can more accurately capture the intended logical structure of the input, aligning with the cognitive processes involved in handling negation.
\section{Pretrain Boolean-aware Attention}
\label{ap:pretrain}
Introducing new parameters in \emph{BoolAttn} poses a risk of overfitting due to the limited size of Boolean retrieval datasets compared to the model's capacity when integrated into transformer layers. This imbalance may lead the model to memorize the training data rather than generalizing effectively, resulting in suboptimal performance. To mitigate this, we pretrain Bool-BERT on \citealp{mai2024setbert}'s GPT-generated dataset on two auxiliary tasks: cue prediction and expert classification (gate selection). The cue position prediction loss is calculated as follows:
\begin{multline*}   
\mathcal{L}_{cue} = -\frac{1}{B\times L}\sum_{i=1}^B\sum_{j=1}^L(y_{i,j}\log p_{i, j}+ \\ (1 - y_{i,j})\log (1 - p_{i, j}))
\end{multline*}
where $y$ is the ground-truth label and $p$ is the predicted probability that token $j$ in batch $i$ is a cue. The original dataset only includes ``and,'' ``or," and ``not" as Boolean cues. To enhance model robustness, we augment the dataset by replacing cues with their semantic equivalents, such as substituting ``and" with ``as well as," ``are also," or ``including," and replacing ``not" with ``other than" or ``excluding." Consequently, multiple cue tokens can appear in a single sequence. We use BERT's tokenizer, followed by a word-matching approach to label the cue tokens. 

The second task in pretraining phase is \emph{Booealn expert classfication}. This task is a multi-label classification problem designed to identify which Boolean operators are present in a given input sequence. Since a sequence may contain more than one Boolean operation, this task requires predicting multiple labels simultaneously (one for each operator). Given $E$ Boolean experts (3 if we use \emph{and}, \emph{or}, \emph{not}), the loss is defined as
\begin{multline*}     
\mathcal{L}_{gate} = -\frac{1}{B\times E}\sum_{i=1}^B\sum_{k=1}^E(y_{i,k}\log p_{i, k}+ \\
(1 - y_{i,k})\log (1 - p_{i, k}))
\end{multline*}
where $y$ is the ground-truth label and $p_{i, k}$ is the predicted probability for operator $k$ in batch $i$.

Until now we only pretrain the CuePred and Gate modules, the parameters for ScopePred and BiasPred remain uninitialized. To provide these modules with a reasonable starting point without overfitting them to Boolean logic, we train them on the Boolean sentence classification task from \citealp{mai2024setbert} using a standard triplet loss. Although \citealp{mai2024setbert} reported that triplet loss hinders performance on Boolean queries, we deliberately use it to provide an "okay" initialization, allowing the modules to learn actively during the main tasks. The pretraining objective is defined as follows:

$$\mathcal{L}_{pretrain} = \mathcal{L}_{cue} + \mathcal{L}_{gate} + \alpha\mathcal{L}_{triplet}$$
where $\alpha$ is set to a small value, we used $0.2$. During pretraining, only the \emph{BoolAttn} modules and BERT's embedding layer are updated, while the other layers are frozen. We train for a single epoch with a low learning rate of $1e-5$.

\section{BoolAttn hyperparameters and settings}
All experiments are conducted on two NVIDIA Quadro RTX 8000 GPUs, each with 50GB of memory. We use PyTorch \cite{paszke2019pytorch} as our deep learning framework and leverage Hugging Face's Transformers library \cite{wolf2020transformers} for pretrained models and model implementation.

\paragraph{Hyperparameters}
(1) Operator embedding dimension $d_h$: 10. (2) Threshold $\theta$ is initialized at 0.5. (3) Relative postion window: 5. (4) Gaussian kernal sigma: $\sigma$: 2.

\paragraph{Gating mechanism}
For the learnable gate mechanism, we employ a simple FFN followed by a sigmoid activation, with logits scaled by a small temperature value of $0.1$ to enhance sensitivity. Unlike softmax, which enforces mutual exclusivity, sigmoid is better suited for Boolean logic since multiple Boolean operators can coexist within a single query. By applying a hard threshold to the sigmoid outputs, the gates can selectively activate multiple Boolean experts, allowing them to contribute to the final attention scores as needed.

The gating mechanism introduces a performance bottleneck as it relies on accurately activating the Boolean experts corresponding to the operators present in the query. This dependency can lead to suboptimal performance (Table~\ref{tab:learnable-gate}). Improving this gating strategy is left for future work. However, in real-world applications, users could explicitly specify the Boolean logic in their queries. This information could then be used as an auxiliary label to activate the relevant Boolean experts—an approach we currently employ.

\begin{table}[]
\centering
\begin{tabular}{lcccc}
\hline
                           & \multicolumn{3}{c}{Avg. Recall@K}                                 \\ \cline{2-4} 
Retriever                            & 50             & 100            & 1000           \\ \hline
Bool-Bert\textit{-base}                          & 0.379          & 0.483          & 0.788          \\
Bool-Bert\textit{-base}-gate         & 0.371          & 0.470          & 0.775 \\
\end{tabular}
\caption{Learnable gate introduces bottleneck.}
\label{tab:learnable-gate}
\end{table}

\section{Dense Retriever}
We adopt a dual-encoder framework \cite{karpukhin2020dense} for our retriever.

Unlike \citealp{malaviya2023quest} and \citealp{mai2024setbert}, which use a fixed positive document per query, we sample different positive documents at each training step. This encourages the encoders to explore diverse relevant documents for the same query, which is crucial for multi-answer retrieval tasks where multiple correct answers exist. This is particularly important for boolean queries: negation excludes a subset of answers, disjunctions (unions) expand the answer set, and conjunctions (intersections) narrow it. By introducing varied positive samples, our approach enhances model robustness and generalization. Furthermore, we empirically found that using random negative samples is more effective than hard negative samples (i.e., incorrect answers retrieved by BM25), particularly for the QUEST dataset. Table~\ref{tab:positive-sample} shows the improvement when using different positive documents and normal negative samples, T5 and BERT results taken from \citealp{malaviya2023quest} and \citealp{mai2024setbert}, respectively.

\begin{table}[]
\centering
\begin{tabular}{lcccc}
\hline
                           & \multicolumn{3}{c}{Avg. Recall@K}                                 \\ \cline{2-4} 
Retriever                            & 50             & 100            & 1000           \\ \hline
BERT-Base                          & 0.134          & 0.182          & 0.453          \\
BERT-Large                          & 0.227          & 0.300          & 0.627          \\
T5-Base                              & 0.372          & 0.455          & 0.726          \\
T5-Large                    & \textbf{0.386} & \textbf{0.476} & 0.757          \\ \hline
BERT-Base-Ours          & 0.344          & 0.443          & \textbf{0.760} \\
\end{tabular}
\caption{Importance of choosing different positive samples, results for QUEST dataset.}
\label{tab:positive-sample}
\end{table}

\paragraph{Hyperparameters}

\begin{enumerate}
    \item Max query length: 64.
    \item Max document length: 256.
    \item Negative document per sample: 5.
    \item Positive document: 1, shuffled every step.
    \item Batch size: 32.
    \item Optimizer: AdamW \cite{kingma2015adam} with weight decay 0.01 \cite{loshchilov2019decoupled}.
    \item Learning rate: 5e-5.
    \item Epoch: 40.
    \item Evaluation every epoch, keep best.
\end{enumerate}

\end{document}